\documentclass{article}
\usepackage{microtype}
\usepackage{graphicx}
\usepackage{subfigure}
\usepackage{booktabs}
\usepackage{hyperref}
\usepackage{xcolor}
\usepackage{fullpage}

\usepackage{amssymb,amsmath,amsthm}
\usepackage{empheq}
\usepackage{algorithm,algorithmic}

\newcommand{\ABound}{\kappa_A}
\newcommand{\BBound}{\kappa_B}
\newcommand{\PertBound}{W}
\newcommand{\Diameter}{D}
\newcommand{\CostBound}{\beta}
\newcommand{\CostGradBound}{G}
\newcommand{\hor}{H}

\newcommand{\BigBound}{D}
\newcommand{\dimn}{d}

\def\M{{\mathcal M}}

\def\K{{\mathcal K}}

\def\reals{{\mathbb R}}

\def\norm#1{\mathopen\| #1 \mathclose\|}

\newcommand{\ignore}[1]{}

\def\reals{{\mathbb R}}

\def\bold0{\mathbf{0}}

\def\epsilon{\varepsilon}

\def\grad{\nabla}
\newcommand{\defeq}{\triangleq}

\newtheorem{theorem}{Theorem}[section]

\newtheorem{lemma}[theorem]{Lemma}

\newtheorem{definition}[theorem]{Definition}

\newtheorem{assumption}[theorem]{Assumption}

\newcommand{\newreptheorem}[2]{%
\newenvironment{rep#1}[1]{%
 \def\rep@title{#2 \ref{##1}}%
 \begin{rep@theorem}}%
 {\end{rep@theorem}}}

\newreptheorem{theorem}{Theorem}
\newreptheorem{lemma}{Lemma}
\newreptheorem{proposition}{Proposition}
\newreptheorem{claim}{Claim}
\newreptheorem{corollary}{Corollary}
\newreptheorem{mainlemma}{Main Lemma}

\newcommand{\namedref}[2]{\mbox{\hyperref[#2]{#1~\ref*{#2}}}}

\newcommand{\sectionref}[1]{\namedref{Section}{#1}}
\newcommand{\appendixref}[1]{\namedref{Appendix}{#1}}

\newcommand{\figurerefb}[2]{\mbox{\hyperref[#1]{Figure~\ref*{#1}#2}}}

\newcommand{\equationref}[1]{\mbox{\hyperref[#1]{(\ref*{#1})}}}
\renewcommand{\eqref}{\equationref}

\numberwithin{equation}{section}

\newcommand{\braces}[1]{\left\{#1\right\}}
\newcommand{\pa}[1]{\left(#1\right)}

\newcommand{\abs}[1]{\left|#1\right|}

\def\memdiam{D}
\def\memgradbound{G_f}
\begin{document}
\title{Online Control with Adversarial Disturbances}
\author{
  Naman Agarwal$^{1}$ \qquad Brian Bullins$^{1\,2}$ \qquad Elad Hazan$^{1\,2}$ \\ Sham M. Kakade$^{1\,3\,4}$ \qquad Karan Singh$^{1\,2}$\\
  \\
  $^1$ Google AI Princeton \\
  $^2$ Department of Computer Science, Princeton University \\
  $^3$ Allen School of Computer Science and Engineering, University of Washington \\
  $^4$ Department of Statistics, University of Washington\\
  \texttt{namanagarwal@google.com}, \texttt{\{bbullins,ehazan,karans\}@princeton.edu},\\\texttt{sham@cs.washington.edu} \\
}\maketitle
\begin{abstract}
We study the control of a linear dynamical system with adversarial disturbances (as opposed to statistical noise). The objective we consider is one of regret: we desire an online control procedure that can do nearly as well as that of a procedure that has full knowledge of the disturbances in hindsight. Our main result is an efficient algorithm that provides nearly tight regret bounds for this problem. From a technical standpoint, this work generalizes upon previous work in two main aspects:  our model allows for adversarial noise in the dynamics, and allows for general convex costs.  
\end{abstract}

\section{Introduction}

This paper studies the robust control of linear dynamical systems. 
A linear dynamical system is governed by the dynamics equation
\begin{equation}\label{eqn:shalom}
    x_{t+1} = Ax_t + Bu_t + w_t,  
\end{equation}
where $x_t$ is the state, $u_t$ is the control and $w_t$ is a disturbance to the system. At every time step $t$, the controller suffers a cost $c(x_t,u_t)$ to enforce the control. In this paper, we consider the setting of online control with \emph{arbitrary} disturbances. Formally, the setting involves, at every time step $t$, an adversary selecting a convex cost function $c_t(x,u)$ and a disturbance $w_t$, and the goal of the controller is to generate a sequence of controls $u_t$ such that a sequence of convex costs $c_t(u_t,x_t)$ is minimized. 

The above setting generalizes a fundamental problem in control theory (including the Linear Quadratic Regulator) which has been studied over several decades, surveyed below. However, despite the significant research literature on the problem, our generalization and results address several challenges that have remained. 

{\bf Challenge 1.} Perhaps the most important challenge we address is in dealing with arbitrary disturbances $w_t$ {\it in the dynamics}. This is a difficult problem, and so standard approaches almost exclusively assume i.i.d. Gaussian noise. Worst-case approaches in the control literature, also known as $H_\infty$-control and its variants, are overly pessimistic. Instead, we take an online (adaptive) approach to dealing with adversarial disturbances. 

{\bf Challenge 2.} Another limitation for efficient methods is the classical assumption that the costs $c(x_t,u_t)$ are quadratic, as is the case for the  {\it linear quadratic regulator}. Part of the focus in the literature on the quadratic costs is due to special properties that allow for efficient computation of the best linear controller in hindsight. One of our main goals is to introduce a more general technique that allows for efficient algorithms even when faced with arbitrary convex costs.

\paragraph{Our contributions.}
In this paper, we tackle both challenges outlined above: coping with adversarial noise, and general loss functions in an online setting. For this we turn to the time-trusted methodology of regret minimization in online learning. In the field of online learning, regret minimization is known to be more robust and general than statistical learning, and a host of convex relaxation techniques are readily available. To define the performance metric, denote for any control algorithm $\mathcal{A}$, 
\[ J_T(\mathcal{A}) = \sum_{t=1}^T c_t(x_t,u_t). \]
The standard comparator in control is a linear controller, which generates a control signal as a linear function of the state, i.e.  $u_t=-Kx_t$. 
Let $J(K)$ denote the cost of a linear controller from a certain class $K \in \K$. 
For an algorithm $\mathcal{A}$, we define the regret as the sub-optimality of its cost with respect to the best linear controller from a certain set
\begin{equation*}
     \texttt{Regret} = J_T(\mathcal{A}) - \min_{K\in \K}J_T(K).
\end{equation*}
Our main result is an efficient algorithm for control which achieves regret $O(\sqrt{T})$ in the setting described above. A similar setting has been considered in literature before \cite{cohen2018online}, but our work generalizes previous work in the following ways:
\begin{enumerate}
    \item Our algorithm achieves regret $O(\sqrt{T})$ even in the presence of bounded adversarial disturbances. Previous regret bounds needed to assume that the  disturbances $w_t$ are drawn from a distribution with zero mean and bounded variance. 
    \item Our regret bounds apply to any sequence of adversarially chosen convex loss functions. Previous efficient algorithms applied to convex quadratic costs only. 
\end{enumerate}

Our results above are obtained using a host of techniques from online learning and online convex optimization, notably online learning for loss functions with memory and improper learning using convex relaxation. 

\section{Related Work}

\paragraph{Online Learning:} Our approach stems from the study of regret minimization in online learning, this paper advocates for worst-case regret as a robust performance metric in the presence of adversarial nosie. A special case of our study is that of regret minimization in stateless (with $A=0$) systems, which is a well studied problem in machine learning. See books and surveys on the subject \cite{cesa2006prediction, OCObook, shalev2012online}.  
Of particular interest to our study is the setting of online learning with memory \cite{anava2015online}.

\paragraph{Learning and Control in Linear Dynamical Systems:} 

The modern setting for linear dynamical systems arose in the seminal work of Kalman \cite{kalman1960new}, who introduced the Kalman filter as a recursive least-squares solution for maximum likelihood estimation (MLE) of Gaussian perturbations to the system in latent-state systems. The framework and filtering algorithm have proven to be a mainstay in control theory and time-series analysis; indeed, the term \emph{Kalman filter model} is often used interchangeably with LDS. We refer the reader to the classic survey \cite{ljung1998system}, and the extensive overview of recent literature in \cite{hardt2016gradient}. 
Most of this literature, as well as most of classical control theory, deals with zero-mean random noise, mostly Normally distributed. 

Recently, there has been a renewed interest in learning both fully-observable \& latent-state linear dynamical systems. Sample complexity and regret bounds (for Gaussian noise) were obtained in \cite{a1,a2}. The fully-observable and convex cases were revisited in \cite{DBLP:journals/corr/abs-1805-09388,simchowitz2018learning}. The technique of spectral filtering for learning and controlling non-observable systems was introduced and studied in \cite{hazan2018spectral,arora2018towards,hazan2017learning}.  Provable control in the Gaussian noise setting was also studied in \cite{fazel2018global}.

\paragraph{Robust Control:} The most notable attempts to handle adversarial perturbations in the dynamics are called $H_\infty$ control \cite{z1,z2}. In this setting, the controller solves for the best linear controller assuming worst case noise to come, i.e. 
$$ \min_{K_1} \max_{\epsilon_{1:T} } \min_{K_2} ... \min_{K_{t}} \max_{\epsilon_{T} } \sum_t c_t(x_t,u_t) , $$
assuming similar linear dynamics as in equation \eqref{eqn:shalom}. In comparison, we do not solve for the entire noise trajectory in advance, but adjust for it iteratively. 
Another difference is computational: the above mathematical program may be hard to compute for general cost functions, as compared to our efficient gradient-based algorithm. 

\paragraph{Non-stochastic MDPs:} The setting we consider, control in systems with linear transition dynamics \cite{bertsekas2005dynamic} in presence of adversarial disturbances, can be cast as that of planning in an adversarially changing MDP \cite{arora2012online, dekel2013better}. The results obtained via this reduction are unsatisfactory because these regret bounds scale with the size of the state space, which is usually exponential in the dimension of the system. In addition, the regret in these scale as $\Omega(T^\frac{2}{3})$. In comparison, \cite{yu2009markov, even2009online} solve the online planning problem for MDPs with fixed dynamics and changing costs. The satisfying aspect of their result is that the regret bound does not explicitly depend on the size of the state space, and scales as $O(\sqrt{T})$. However, the dynamics are fixed and without (adversarial) noise.  

\paragraph{LQR with changing costs:} For the Linear Quadratic Regulator problem, \cite{cohen2018online} consider changing quadratic costs with stochastic noise to get a $O(\sqrt{T})$ regret bound. This work is well aligned with results, and the present paper employs some notions developed therein (eg. strong stability). However, the techniques used in \cite{cohen2018online} (eg. the SDP formulation for a linear controller)  are strongly reliant on the quadratic nature of the cost functions and stochasticity of the disturbances. In particular, even for the offline problem, to the best of our knowledge, there does not exist a SDP formulation to determine the best linear controller for convex losses. In an earlier work, \cite{abbasi2011regret} considers a more restricted setting with fixed, deterministic dynamics (hence, noiseless) and changing quadratic costs.

\section{Problem Setting}
\subsection{Interaction Model}

The Linear Dynamical System is a Markov decision process on continuous state and action spaces, with linear transition dynamics. In each round $t$, the learner outputs an action $u_t$ on observing the state $x_t$ and incurs a cost of $c_t(x_t, u_t)$, where $c_t(\cdot,\cdot)$ is convex. The system then transitions to a new state $x_{t+1}$ according to
\[ x_{t+1} = Ax_t + Bu_t + w_t. \]
In the above definition, $w_t$ is the disturbance sequence the system suffers at each time step. In this paper, we make no distributional assumptions on $w_t$.  The sequence $w_t$ is not made known to the learner in advance.

For any algorithm $\mathcal{A}$, the cost we attribute to it is
\begin{equation*}
J_T(\mathcal{A}) = \sum_{t=1}^T c_t(x_t,u_t)
\end{equation*}
where $x_{t+1}=Ax_t+Bu_t+w_t$ and $u_t=\mathcal{A}(x_1, \dots x_t)$. With some abuse of notation, we shall use $J(K)$ to denote the cost of a linear controller $\pi(K)$ which chooses the action as $u_t=-Kx_t$.

\subsection{Assumptions}

We make the following assumptions throughout the paper. We remark that they are less restrictive, and hence, allow for more general systems than those considered by the previous works. In particular, we allow for adversarial (rather than i.i.d. stochastic) noise, and convex cost functions. Also, the non-stochastic nature of the disturbances permits, without loss of generality, the assumption that $x_0=0$.

\begin{assumption}\label{a1}
The matrices that govern the dynamics are bounded, ie., $\|A\|\leq\ABound, \|B\|\leq\BBound$. The perturbation introduced per time step is bounded, ie., $\|w_t\|\leq \PertBound$. 
\end{assumption}

\begin{assumption}\label{a2}
The costs $c_t(x, u)$ are convex. Further, as long as it is guaranteed that $\|x\|, \|u\|\leq \Diameter$, it holds that
\[ |c_t(x, u)|\leq \CostBound \Diameter^2, \|\nabla_x c_t(x,u)\|, \|\nabla_u c_t(x,u)\|\leq \CostGradBound\Diameter.\]
\end{assumption}

Following the definitions in \cite{cohen2018online}, we work on the following class of linear controllers.

\begin{definition}
A linear policy $K$ is $(\kappa, \gamma)$-strongly stable if there exist matrices $L,H$ satisfying  $A-BK=HLH^{-1}$, such that following two conditions are met:
\begin{enumerate}
    \item The spectral norm of $L$ is strictly smaller than unity, ie., $\|L\| \leq 1-\gamma$.
    \item The controller and the transforming matrices are bounded, ie., $\|K\|\leq \kappa$ and $ \|H\|,\|H^{-1}\| \leq \kappa$.
\end{enumerate}
\end{definition}

\subsection{Regret Formulation}
Let $\K=\{K: K \text{ is } (\kappa,\gamma)\text{-strongly stable}\}$. For an algorithm $\mathcal{A}$, the regret is the sub-optimality of its cost with respect to a best linear controller.
\begin{align*}
    & \texttt{Regret} = J_T(\mathcal{A}) - \min_{K\in \K}J_T(K).
\end{align*}

\subsection{Proof Techniques and Overview}

\paragraph{Choice of Policy Class:}We begin by parameterizing the policy we execute at every step as a linear function of the disturbances in the past in Definition \ref{defn:policy}. Similar parameterization has been considered in the system level synthesis framework (see \cite{wang2019system}). This leads to a convex relaxation of the problem. Optimization on alternative paramterizations including an SDP based framework \cite{cohen2018online} or a direct parametrization \cite{fazel2018global} have been studied in literature but they seem unable to capture general convex functions as well as adversarial disturbance or lead to a non-convex loss. To avoid a linear dependence on time for the number of parameters in our policy we additionally include a stable linear controller in our policy allowing us to effectively consider only $O(\gamma^{-1}\log(T))$ previous perturbations. Lemma \ref{l:repstat} makes this notion of approximation precise. 

\paragraph{Reduction to OCO with memory:} The choice of the policy class with an appropriately chosen horizon $\hor$ allows us to reduce the problem to compete with functions with truncated memory. This naturally falls under the class of online convex optimization with memory (see Section \ref{sec:oco_mem}). Theorem \ref{thm:approxthm} makes this reduction precise. Finally to bound the regret on truncated functions we use the Online Gradient Descent based approach specified in \cite{anava2015online}, which requires a bound on Lipschitz constants which we provide in Section \ref{sec:const_bounds}. This reduction is inspired from the ideas introduced in \cite{even2009online}.     

\subsection{Roadmap}

The next section provides the suite of definition and notation required to define our algorithm and regret bounds. Section \ref{sec:main} contains our main algorithm \ref{alg:mainA} and the main regret guarantees \ref{thm:main} followed by the proof and the requisite lemmas and their respective proofs.  
\section{Preliminaries}
In this section, we establish some important definitions that will prove useful throughout the paper.
\subsection{Notation}

We reserve the letters $x,y$ for states and $u,v$ for control actions. 
We denote by $\dimn = \max(\mathrm{dim}(x), \mathrm{dim}(u))$, i.e., a bound on the dimensionality of the problem. We reserve capital letters $A,B,K,M$ for matrices associated with the system and the policy. Other capital letters are reserved for universal constants in the paper.

\subsection{A Disturbance-Action Policy Class}
We put forth the notion of a \emph{disturbance-action controller} which chooses the action as a linear map of the past disturbances. Any disturbance-action controller ensures that the state of a system executing such a policy may be expressed as a linear function of the parameters of the policy. This property is convenient in that it permits efficient optimization over the parameters of such a policy. The situation may be contrasted with that of a linear controller. While the action recommended by a linear controller is also linear in past disturbances (a consequence of being linear in the current state), the state sequence produced on the execution of a linear policy is a not a linear function of its parameters.

\begin{definition}[Disturbance-Action Policy]\label{defn:policy}
A disturbance-action policy $\pi(M,K)$
is specified by parameters $M=(M^{[1]},\dots,M^{[\hor]})$ and a fixed matrix $K$. At every time $t$, such a policy $\pi(M,K)$ chooses the recommended action $u_{t}$ at a state $x_t$\footnote{$x_t$ is completely determined given $w_0 \ldots w_{t-1}$. Hence, the use of $x_t$ only serves to ease presentation.}, defined as
\[ u_t = -Kx_t+\sum_{i=1}^{\hor} M^i w_{t-i}. \]
For notational convenience, here it may be considered that $w_i=0$ for all $i<0$.
\end{definition}

We refer to the policy played at time $t$ as $M_t = \{M_t^{[i]}\}$ where the subscript $t$ refers to the time index and the superscript $[i]$ refers to the action of $M_t$ on $w_{t-i}$. Note that such a policy can be executed because $w_{t-1}$ is perfectly determined on the specification of $x_{t}$ as $w_{t-1}=x_{t}-Ax_{t-1}-Bu_{t-1}$. It shall be established in later sections that such a policy class can approximate any linear policy with a strongly stable matrix in terms of the total cost suffered.

\subsection{Evolution of State}
In this section, we reason about the evolution of the state of a linear dynamical system under a non-stationary policy $\pi=(\pi_0,\dots,\pi_{T-1})$ composed of $T$ policies, where each $\pi_t$ is specified by $\pi_t(M_t=(M_t^{[1]},\dots,M_t^{[\hor]}),K)$. Again, with some abuse of notation, we shall use $\pi((M_0,\dots,M_{T-1}), K)$ to denote such a non-stationary policy.

The following definitions serve to ease the burden of notation.
\begin{enumerate}
    \item Define $\tilde{A}_K=A-BK$. $\tilde{A}_K$ shall be helpful in describing the evolution of state starting from a non-zero state in the absence of disturbances.
    \item $x^K_t(M_0,\dots,M_{t-1})$ is the state attained by the system upon execution of a non-stationary policy $\pi((M_0,\dots,M_{t-1}), K)$. We drop the arguments $M_i$ and the $K$ from the definition of $x_t$ when it is clear from the context. If the same policy $M$ is used across all time steps, we compress the notation to $x_t^K(M)$. Note that $x_t^{K}(0)$ refers to running the linear policy $K$ in the standard way. 
    \item $\Psi_{t,i}^K(M_0\dots,M_{t-1})$ is a transfer matrix that describes the effect of $w_{t-i}$ on the state $x_{t+1}$, formally defined below. When the arguments to $\Psi_{t,i}^K$ are clear from the context, we drop the arguments. When $M$ is the same across all arguments we suppress the notation to $\Psi_{t,i}^K(M)$.
\end{enumerate}

\begin{definition}
Define the disturbance-state transfer matrix $\Psi_{t,i}^K$ to be
\begin{align*}
    \Psi_{t,i}^K(M_{t-H},\dots,M_{t-1}) = \tilde{A}_K^i \mathbf{1}_{i\leq H} + \sum_{j=1}^{\hor} \tilde{A}_K^j BM_{t-j}^{[i-j]} \mathbf{1}_{i-j \in [1,\hor]}.
\end{align*}
\end{definition}


It will be worthwhile to note that $\Psi_{t,i}^K$ is linear in $M_{t-1},\dots,M_{t-H}$.

\begin{lemma}
If $u_t$ is chosen as a non-stationary policy $\pi((M_1,\dots,M_T), K)$ recommends, then the state sequence is governed as follows:
\begin{equation}
    x_{t+1} = \sum_{i=0}^t \Psi_{t,i} w_{t-i},
\end{equation}
which can equivalently be written as
\begin{equation}
     x_{t+1} = \tilde{A}_K^{\hor+1}x_{t-\hor}+\sum_{i=0}^{2\hor} \Psi_{t,i} w_{t-i}.
\end{equation}
\end{lemma}

\subsection{Idealized Setting}

Note that the counter-factual nature of regret in the control setting implies in the loss at a time step $t$, depends on all the choices made in the past. To efficiently deal with this  we propose that our optimization problem only consider the  effect of the past $\hor$ steps while planning, forgetting about the state, the system was at time $t - \hor$. We will show later that the above scheme tracks the true cost suffered upto a small additional loss. To formally define this idea, we need the following definition on \emph{ideal} state.

\begin{definition}[Ideal State \& Action]
Define an \emph{ideal} state $y_{t+1}^K$ which is the state the system would have reached if it played the non-stationary policy $(M_{t-\hor},\dots,M_t)$ at all time steps from $t-\hor$ to $t$, assuming the state at $t-\hor$ is $0$. Similarly, define $v_t^K(M_{t-\hor}, \dots,M_t)$ to be an idealized action that would have been executed at time $t$ if the state observed at time $t$ is $y_t^K(M_{t-\hor},\dots, M_{t-1})$. Formally,
\begin{align*}
y^K_{t+1}(M_{t-\hor},\dots,M_{t}) &= \sum_{i=0}^{2\hor} \Psi_{t,i} w_{t-i}, \\
v_{t}^K(M_{t-\hor}, \dots,M_{t}) &= -Ky^K_t+\sum_{i=1}^\hor M_t^{[i]} w_{t-i}.
\end{align*}
\end{definition}
We can now consider the loss of the \emph{ideal} state and the \emph{ideal} action. 

\begin{definition}[Ideal Cost]
\label{defn:idealcost}
Define the idealized cost function $f_t$ to be the cost associated with the idealized state and idealized action, i.e.,
\begin{align*}
    f_{t}(M_{t-\hor}, \dots,M_{t}) = c_t(y^K_{t}(M_{t-\hor}, \dots,M_{t-1}), v^K_{t}(M_{t-\hor}, \dots,M_{t})).
\end{align*}

\end{definition}

The linearity of $y_t^K$ in past controllers and the linearity of $v_t^K$ in its immediate state implies that $f_t$ is a convex function of a linear transformation of $M_{t-\hor}, \dots,M_t$ and hence convex in $M_{t-\hor}, \dots,M_t$. This renders it amenable to algorithms for online convex optimization. 

In Theorem \ref{thm:approxthm} we show that $f_t$ and $c_t$ on a sequence are close by and this reduction allows us to only consider the truncated $f_t$ while planning allowing for efficiency. The precise notion of minimizing regret such truncated $f_t$ was considered in online learning literature \cite{anava2015online} before as online convex optimization(OCO) with memory. We present an overview of this framework next.   

\subsection{OCO with Memory}
\label{sec:oco_mem}
We now present an overview of the online convex optimization (OCO) with memory framework, as established by \cite{anava2015online}. In particular, we consider the setting where, for every $t$, an online player chooses some point $x_t \in \K \subset \reals^d$, a loss function $f_t : \K^{\hor+1}\mapsto \reals$ is revealed, and the learner suffers a loss of $f_t(x_{t-\hor},\dots,x_t)$. We assume a certain coordinate-wise Lipschitz regularity on $f_t$ of the form such that, for any $j \in \braces{1,\dots,\hor}$, for any $x_1, \dots, x_{\hor},\tilde{x}_j\in \K$,
\begin{align}\label{eq:memlip}
    \abs{f_t(x_1,\dots,x_j,\dots,x_{\hor}) - f_t(x_1,\dots,\tilde{x}_j,\dots,x_\hor)} \leq L\norm{x_j - \tilde{x}_j}.
\end{align}
In addition, we define $\tilde{f}_t(x) = f_t(x,\dots,x)$, and we let
\begin{equation}\label{eq:membounds}
    \memgradbound = \sup\limits_{t \in \braces{1,\dots,T}, x \in \K} \norm{\grad \tilde{f}_t(x)}, \; \text{  }\;  \memdiam = \sup\limits_{x,y\in \K} \norm{x-y}.
\end{equation}

The resulting goal is to minimize the \emph{policy regret} \cite{arora2012online}, which is defined as
\begin{align*}
    \texttt{PolicyRegret} = \sum\limits_{t=\hor}^T f_t(x_{t-\hor},\dots,x_t) - \min\limits_{x \in \K} \sum\limits_{t=\hor}^T f_t(x,\dots,x).
\end{align*}

As shown by \cite{anava2015online}, by running a memory-based OGD, we may bound the policy regret by the following theorem.
\begin{theorem}\label{thm:oco_memory}
Let $\braces{f_t}_{t=1}^T$ be Lipschitz continuous loss functions with memory such that $\tilde{f}_t$ are convex, and let $L$, $\memdiam$, and $\memgradbound$ be as defined in \eqref{eq:memlip} and \eqref{eq:membounds}. Then, Algorithm \ref{ogdm} generates a sequence $\braces{x_t}_{t=1}^T$ such that
\begin{align*}
  \sum\limits_{t=\hor}^T f_t(x_{t-\hor},\dots,x_t) - \min\limits_{x \in \K} \sum\limits_{t=\hor}^T f_t(x,\dots,x) \leq \frac{\memdiam^2}{\eta} + T\memgradbound^2\eta + L\hor^2\eta \memgradbound T.   
\end{align*}

Furthermore, setting $\eta = \frac{D}{\sqrt{\memgradbound(\memgradbound+LH^2)T}}$ implies that \begin{equation*}
    \texttt{\emph{PolicyRegret}} \leq O\pa{D\sqrt{\memgradbound(\memgradbound+L\hor^2)T}}.
\end{equation*}
\end{theorem}


\section{Algorithm \& Main Result}
\label{sec:main}
Algorithm \ref{alg:mainA} describes our proposed algorithm for controlling linear dynamical systems with adversarial disturbances which at all times maintains a disturbance-action controller. The algorithm implements the memory based OGD on the loss $f_t(\cdot)$ as described in the previous section. The algorithm requires the specification of a $(\kappa,\gamma)$-strongly stable matrix $K$ once before the online game. Such a matrix can be obtained offline using an SDP relaxation as described in \cite{cohen2018online}. The following theorem states the regret bound Algorithm \ref{alg:mainA} guarantees.

\begin{algorithm}[t!]
\caption{Online Control Algorithm}
\label{alg:mainA}
\begin{algorithmic}[1]
\STATE \textbf{Input:} Step size $\eta$, Control Matrix $K$, Parameters $\BBound, \kappa, \gamma, T$.
\STATE Define $\hor = 2\BBound \kappa^3 \gamma^{-1} \log(T)$
\STATE Define $\M = \{M = \{M^{[1]} \ldots M^{[\hor]}\}: \|M^{[i]}\| \leq \kappa^3 \BBound (1 - \gamma)^i\}$.
\STATE Initialize $M_0\in \M$ arbitrarily.
\FOR{$t = 0, \ldots, T-1$}
\STATE Choose the action $u_t = c_t-Kx_t+\sum_{i=1}^{\hor} M^{[i]} w_{t-i}$.
\STATE Observe the new state $x_{t+1}$ and record $w_t=x_{t+1}-Ax_t-Bu_t$.
\STATE Define the function $g_t(M)$ as
$g_t(M) = f_t(M, \dots M)$ (refer Definition \ref{defn:idealcost})
\STATE Set $M_{t+1} = \Pi_{\M}(M_t - \eta \nabla g_{t}(M))$ 
\ENDFOR
\end{algorithmic}
\end{algorithm}

\begin{theorem}[Main Theorem]\label{thm:main}
Suppose Algorithm~\ref{alg:mainA} is executed with $\eta=\Theta\left(\CostGradBound\PertBound\sqrt{T}\right)^{-1}$, on an LDS satisfying Assumption~\ref{a1} with control costs satisfying  Assumption ~\ref{a2}. Then, it holds true that
\[J_T(\mathcal{A}) - \min_{K\in \K}J_T(K) \leq O\left(\CostGradBound \PertBound^2 \sqrt{T} \log(T)\right),\]
Furthermore, the algorithm maintains at most $O(1)$ parameters can be implemented in time $O(1)$ per time step. Here $O(\cdot)$, $\Theta(\cdot)$ contain polynomial factors in  $\gamma^{-1}, \BBound, \kappa, \dimn$. 
\end{theorem}
\begin{proof}[Proof of Theorem \ref{thm:main}]
Note that by the definition of the algorithm we have that all $M_t \in \M$, where 
\[ \M = \{M = \{M^{[1]} \ldots M^{[\hor]}\}: \|M^{[i]}\| \leq \kappa^3 \BBound (1 - \gamma)^i\}.\]
Let $\BigBound$ be defined as 
\[\BigBound \triangleq \frac{W(\kappa^2 + \hor\BBound\kappa^2a)}{\gamma ( 1 - \kappa^2(1 - \gamma)^{\hor+1})} + \frac{\BBound \kappa^3 W}{\gamma}.\]

Let $K^*$ be the optimal linear policy in hindsight. By definition $K^*$ is a $(\kappa,\gamma)$-strongly stable matrix. Using Lemma \ref{l:repstat} and Theorem \ref{thm:approxthm}, we have that
\begin{align}
\label{eqn:approxopt}
    &\min_{M_* \in \M} \left(\sum_{t=0}^{T}  f_t(M_*, \ldots, M_*)\right)  - \sum_{t=0}^Tc_t(x_t^{K^*}(0),u_t^{K^*}(0)) \\
    &\leq \min_{M_* \in \M} \left( \sum_{t=0}^T c_t(x_t^K(M_*),u_t^K(M_*))\right) - \sum_{t=0}^Tc_t(x_t^{K^*}(0),u_t^{K^*}(0)) + 2T \CostGradBound \BigBound^2 \kappa^3 (1 - \gamma)^{\hor + 1} \nonumber \\
    &\leq 2T\CostGradBound\BigBound (1 - \gamma)^{\hor + 1}\left(  \frac{\PertBound \hor \kappa_B^2 \kappa^5}{\gamma} + \BigBound \kappa^3 \right).
 \end{align}
Let $M_1 \ldots M_T$ be the sequence of policies played by the algorithm. Note that by definition of the constraint set $S$, we have that
\[ \forall t \in [T], \forall i \in [\hor] \;\;\; \|M_t^{[i]}\| \leq \BBound\kappa^3(1 - \gamma)^i.\]
Using Theorem \ref{thm:approxthm} we have that
\begin{align}
\label{eqn:approxplayed}
    \sum_{t=0}^{T} c_t(x_t^{K},u_t^{K}) - \sum_{t=0}^{T} f_t(M_{t - \hor} \ldots M_t) \leq  2T \CostGradBound \BigBound^2 \kappa^3 (1 - \gamma)^{\hor + 1}.
\end{align}

Finally using Theorem \ref{thm:oco_memory} and using Lemmas \ref{lemma:memory_lipsconst}, \ref{lemma:biggradbound} to bound the constants $\memgradbound$ and $L$ associated with the function $f_t$ and by noting that 
\[\max_{M_1, M_2 \in \M} \|M_1 - M_2\| \leq \frac{\BBound \kappa^3\sqrt{d}}{\gamma},\]
we have that 
\begin{align}
    \label{eqn:final}
    \sum_{t=0}^T f_t(M_{t-\hor} \ldots M_t) - \min_{M_* \in \M}\sum_{t=0}^T f_t(M_*, \ldots, M_*) \leq 8\CostGradBound \PertBound\BigBound \dimn^{3/2}\BBound^2 \kappa^6 H^{2.5} \gamma^{-1}\sqrt{T}.
\end{align}

Summing up \eqref{eqn:approxopt}, \eqref{eqn:approxplayed} and \eqref{eqn:final}, and using the condition that $\hor = \frac{\kappa^2}{\gamma}\log(T)$, we get the result.\qedhere
\end{proof}

\subsection{Sufficiency of Disturbance-Action Policies}
The class of policies described in Definition~\ref{defn:policy} is powerful enough in its representational capacity to capture any fixed linear policy. Lemma~\ref{l:repstat} establishes this equivalence in terms of the state and action sequence each policy produces.


\begin{lemma}[Sufficiency]\label{l:repstat}
For any two $(\kappa,\gamma)$-strongly stable matrices $K^*, K$, there exists a policy $\pi(M_*, K)$, with $M_*=(M_*^{[1]}, \ldots, M_*^{[\hor]})$ defined as \[M_*^{[i]} = (K^* - K)(A - BK^*)^{i-1}\] such that \begin{align}
    \sum_{t=0}^{T} \left( c_t(x_t^K(M_*),u_t^K(M_*)) - c_t(x_t^{K^*}(0),u_t^{K^*}(0)) \right)  \leq T \cdot \frac{2\CostGradBound\BigBound\PertBound \hor \kappa_B^2 \kappa^5(1 - \gamma)^{\hor + 1}}{\gamma}
\end{align}
\end{lemma}
\begin{proof}[Proof of Lemma \ref{l:repstat}]

By definition we have that 
\[x_{t+1}(K^*) = \sum_{i = 0}^t \tilde{A}_K^iw_{t-i}\]
Consider the following calculation for  $M_*$ with $M_*^{[i]} \defeq (K^* - K)(A - BK^*)^{i-1}$ and for any $i \in \{0 \ldots \hor\}$. We have that 
\begin{align*}
	\Psi_{t,i}^K(M_*) &= \tilde{A}_K^i + \sum_{j=1}^{i}\tilde{A}_K^{i-j}BM^{[j]} \\
	&= \tilde{A}_K^i + \sum_{j=1}^{i}\tilde{A}_K^{i-j}B(K^* - K)\tilde{A}_{K^*}^{j-1} \\
	&= \tilde{A}_K^i + \sum_{j=1}^{i}\tilde{A}_K^{i-j}(\tilde{A}_{K^*} - \tilde{A}_{K})\tilde{A}_{K^*}^{j-1} \\
	&= \tilde{A}_K^i + \sum_{j=1}^{i}\left( \tilde{A}_K^{i-j}\tilde{A}_{K^*}^j - \tilde{A}_K^{i-j+1}\tilde{A}_{K^*}^{j-1} \right) \\
	&= \tilde{A}_{K^*}^i
\end{align*}
The final equality follows as the sum telescopes. Therefore, we have that
\[ x_{t+1}^K(M_*) = \sum_{i=0}^{\hor}\tilde{A}_{K^*}^iw_{t-i} + \sum_{i = \hor+1}^{t} \Psi^K_{t,i}(M_*)w_{t-i}.\]
From the above we get that 
\begin{align}
    \|x_{t}^{K^*}(0) -  x_{t}^K(M_*)\| \leq \PertBound \sum_{i=\hor +1}^t \|\Psi^K_{t,i}(M_*)\| \leq \frac{\PertBound \hor \kappa_B^2 \kappa^5(1 - \gamma)^{\hor + 1}}{\gamma},
\end{align}

where the last inequality follows from using Lemma \ref{lemma:affopnorm} and using the fact that $\|M_*^{[i]}\| \leq \BBound\kappa^3(1-\gamma)^i$.

Further comparing the actions taken by the two policies we get that
\begin{align*}
\|u_t^{K^*} - u_t^K(M_*)\| &= \left\|-K^*x_t^{K^*} + Kx_t^K(M_*) - \sum_{i=0}^t (K^*-K)\tilde{A}_{K^*}^iw_{t-i}\right\| \\
&\leq \left\| \sum_{i=\hor+1}^t K\left(\tilde{A}_{K^*}^i +  \Psi_{t,i}^K(M_*)\right)w_{t-i} \right\|\\
&\leq \frac{2\PertBound \hor \kappa_B^2 \kappa^5(1 - \gamma)^{\hor + 1}}{\gamma}.
\end{align*}
Using the above, Assumption \ref{a2} and Lemma \ref{lemma:statebounds}, we get that
\begin{align}
    \sum_{t=0}^{T} \left( c_t(x_t^K(M_*),u_t^K(M_*)) - c_t(x_t^{K^*},u_t^{K^*}) \right) \leq  T \cdot \frac{2\CostGradBound\BigBound\PertBound \hor \kappa_B^2 \kappa^5(1 - \gamma)^{\hor + 1}}{\gamma}.\qedhere
\end{align}
\end{proof}

\subsection{Approximation Theorems}
The following theorem relates the cost of $f_t(M_{t-\hor}, \ldots M_t)$ with the actual cost $c_t(x_t,u_t)$.

\begin{theorem}
\label{thm:approxthm}
For any $(\kappa, \gamma)$-strongly stable $K$, any number $a$ and any sequence of policies $M_1 \ldots M_T$ satisfying $\|M_t^{[i]}\| \leq a(1 - \gamma)^i$, if the perturbations are bounded by $\PertBound$, we have that
\begin{align}
    \sum_{t=1}^T f_{t}(M_{t - \hor}, \ldots M_t) - \sum_{t=1}^T c_t(x_t^K,u_t^K)  \leq 2T \CostGradBound \BigBound^2 \kappa^3 (1 - \gamma)^{\hor + 1}
\end{align}
where
\[\BigBound \triangleq \frac{W(\kappa^2 + \hor\BBound\kappa^2a)}{\gamma ( 1 - \kappa^2(1 - \gamma)^{\hor+1})} + \frac{aW}{\gamma}\]
\end{theorem}

Before giving the proof of the above theorem, we will need a few lemmas which will be useful. 

\begin{lemma}\label{lemma:affopnorm}
    Let $K$ be a $(\kappa,\gamma)$-strongly stable matrix, $a$ be any number and $M_t$ be a sequence such that for all $i,t$, we have $\|M_t^{[i]}\| \leq a(1 - \gamma)^i$, then we have that for all $i,t$
    \[\|\Psi_{t,i}^K\| \leq \kappa^2 (1 - \gamma)^i\cdot\mathbf{1}_{i \leq \hor} + \hor \BBound \kappa^2 a(1 - \gamma)^i\]
\end{lemma}

\begin{proof}[Proof of Lemma \ref{lemma:affopnorm}]

The proof follows by noticing that \begin{align*}
   \|\Psi_{t,i}^K\| &\leq \|\tilde{A}_K^i\|\mathbf{1}_{i \leq \hor} + \sum_{j=1}^\hor \|\tilde{A}_K^j \|\|B\|\|M_{t-j}^{[i-j]}\| \mathbf{1}_{i-j \in [1, \hor]} \\
   &\leq \kappa^2 (1 - \gamma)^i\cdot\mathbf{1}_{i \leq \hor} + \sum_{j=1}^{\hor} \BBound \kappa^2 a (1 - \gamma)^i \\
   &\leq  \kappa^2 (1 - \gamma)^i\cdot\mathbf{1}_{i \leq \hor} +  \hor \BBound  \kappa^2 a (1 - \gamma)^i,
\end{align*}
where the second and the third inequalities follow by using the fact that $K$ is a $(\kappa, \gamma)$-strongly stable matrix and the conditions on the spectral norm of $M$.
\end{proof}
We now derive a bound on the norm of each of the states.
\begin{lemma}
\label{lemma:statebounds}
Suppose the system satisfies Assumption \ref{a1} and let $M_t$ be a sequence such that for all $i,t$, we have that $ \|M_t^{[i]}\| \leq a(1 - \gamma)^i$ for a number $a$. Define 
\[ \BigBound \triangleq \frac{W(\kappa^2 + \hor\BBound\kappa^2a)}{\gamma ( 1 - \kappa^2(1 - \gamma)^{\hor+1})} + \frac{aW}{\gamma}\]
Further suppose $K^*$ is a $(\kappa,\gamma)$-strongly stable matrix. We have that for all $t$ 
\[\max(\|x_t^K\|,\|y_{t}^K(M_{t - \hor - 1} \ldots M_{t - 1})\|,\|x_t(K^*)\|) \leq \BigBound\]
\[\max(\|u_t^K\|, \|v_t^K(M_{t - \hor} \ldots M_t)\|) \leq \BigBound \]
\[\|x_t^K - y_{t}^K(M_{t - \hor - 1} \ldots M_{t - 1})\| \leq \kappa^2(1 - \gamma)^{\hor+1}\BigBound\]
\[\|u_t^K - v_t^K(M_{t - \hor} \ldots M_t)\| \leq \kappa^3(1 - \gamma)^{\hor+1}\BigBound\]
\end{lemma}

\begin{proof}[Proof of Lemma \ref{lemma:statebounds}]
Using the definition of $x_t$ we have that 
\begin{align*}
    \|x_t^K\| &\leq \kappa^2(1 - \gamma)^{\hor + 1}\|x_{t - \hor}\| + \PertBound \cdot \left(  \sum_{i=0}^{2\hor}\|\Psi_{t,i}\|\right) \\
    &\leq 
 \kappa^2(1 - \gamma)^{\hor + 1}\|x_{t - \hor}\| + \PertBound \cdot \left(  \frac{\kappa^2 + \hor\BBound\kappa^2a}{\gamma}\right)
\end{align*}
 The above recurrence can be seen to easily satisfy the following upper bound.
 \begin{equation}
 \label{eqn:intermed1}
     \|x_t^K\| \leq \frac{W(\kappa^2 + \hor\BBound\kappa^2a)}{\gamma ( 1 - \kappa^2(1 - \gamma)^{\hor+1})} \leq \BigBound
 \end{equation}

A similar bound can easily be established for 

\begin{equation}
\label{eqn:intermed2}
   \|y_{t}^K(M_{t - \hor - 1} \ldots M_{t - 1})\| \leq \PertBound \cdot \left(  \frac{\kappa^2 + \hor\BBound\kappa^2a}{\gamma}\right) \leq \BigBound 
\end{equation}

It is also easy to see via the definitions that 
\begin{align}
\label{eqn:intermed3}
    \|x_t^K - y_{t}^K(M_{t - \hor - 1} \ldots M_{t - 1})\| \leq \|\tilde{A}_K^i\|\|x_{t-\hor}\| \leq \kappa^2(1 - \gamma)^{\hor + 1}\BigBound\;\;\;\;
\end{align}

We can finally bound 
\[ \|x_t^{K^*}(0)\| \leq \frac{\PertBound \kappa^2}{\gamma} \leq \BigBound\]
For the actions we can use the definitions to bound the actions as follows using \eqref{eqn:intermed1} and \eqref{eqn:intermed2}
\[ \|u_t^{K}\| \leq \|Kx_t\| + \sum_{i=1}^H \|M_t^{[i]} w_{t-i}\| \leq \kappa \|x_t^K\| + \frac{a\PertBound}{\gamma} \leq \BigBound\]
\begin{align*}
    \|v_t^K(M_{t - \hor} \ldots M_t)\| \leq  \|Ky_t^K(M_{t - \hor - 1} \ldots M_{t-1})\| + \sum_{i=1}^H \|M_t^{[i]} w_{t-i}\| \leq \BigBound.
\end{align*}
We also have that using \eqref{eqn:intermed3}
\begin{align*}
    &\|u_t^K - v_t^K(M_{t - \hor} \ldots M)\| \\
    &= K(x_t^K - y_{t}^K(M_{t - \hor - 1} \ldots M_{t - 1})) \\ 
    &\leq \kappa^3(1 - \gamma)^{\hor + 1}\BigBound.\qedhere
\end{align*}
\end{proof}
Finally, we prove Theroem \ref{thm:approxthm}.
\begin{proof}[Proof of Theorem \ref{thm:approxthm}]

Using the above lemmas we can now bound the approximation error between $f_t$ and $c_t$ using Assumption \ref{a2} 
\begin{align*}
    &|c_t(x_t, u_t) - f_t(M_{t-\hor} \ldots M_t)| \\
    &= |c_t(x_t, u_t) 
    - c_t(y^K_{t}(M_{t-H-1}, \dots M_{t-1}), v^K_{t}(M_{t-H}, \dots M_{t}))|\\
    &\leq \CostGradBound \BigBound\|x_t - y^K_{t}(M_{t-H-1}, \dots M_{t-1})\| + 
    \CostGradBound \BigBound \|u_t - v^K_{t}(M_{t-H}, \dots M_{t}))\|\\
    &\leq 2\CostGradBound \BigBound^2 \kappa^3 (1 - \gamma)^{\hor + 1}.
\end{align*}
This finishes the proof of Theorem \ref{thm:approxthm}.
\end{proof}

\subsection{Bounding the properties of the OCO game with Memory}

\subsubsection{Bounding the Lipschitz Constant}
\label{sec:const_bounds}
\begin{lemma}
\label{lemma:memory_lipsconst}
Consider two policy sequences $\{M_{t - \hor} \ldots M_{t - k} \ldots M_t\}$ and $\{M_{t - \hor} \ldots \tilde{M}_{t-k} \ldots M_t\}$ which differ in exactly one policy played at a time step $t - k$ for $k \in \{0,\ldots,\hor\}$. Then we have that
\begin{align*}
    |f_t(M_{t - \hor} \ldots M_{t - k} \ldots M_t) - f_t(M_{t - \hor} \ldots \tilde{M}_{t - k} \ldots M_t)| 
    \leq 2\CostGradBound\BigBound\PertBound\kappa_B \kappa^3 (1 - \gamma)^k \sum_{i=0}^{\hor}\left(\|M_{t-k}^{[i]} - \tilde{M}_{t-k}^{[i]}\|\right).
\end{align*}
\end{lemma}
\begin{proof}[Proof of Lemma \ref{lemma:memory_lipsconst}]
For the rest of the proof, we will denote $y_{t+1}^K(\{M_{t - \hor} \ldots M_{t - k} \ldots M_t\})$ as $y_{t+1}^K$ and $y_{t+1}^K(\{M_{t - \hor} \ldots \tilde{M}_{t - k} \ldots M_t\})$ as $\tilde{y}_{t+1}^K$. Similarly define $v_t^K$ and $\tilde{v}_t^K$. It follows immediately from the definitions that
\begin{align*}
    \|y_t^K - \tilde{y}^K_t\| &= \|\tilde{A}_K^k B \sum_{i = 0}^{2\hor}\left( M_{t-k}^{[i-k]} - \tilde{M}_{t-k}^{[i-k]} \right)w_{t-i}\mathbf{1}_{i-k \in [1,\hor]}\| \\
    &\leq \kappa_B \kappa^2 (1 - \gamma)^k\PertBound \sum_{i=0}^{\hor}\left(\|M_{t-k}^{[i]} - \tilde{M}_{t-k}^{[i]}\|\right).
\end{align*}
Furthermore, we have that
\begin{align*}
    \|v_t^K - \tilde{v}_t^K\| &= \| - K(y_t - \tilde{y}_t) + \mathbf{1}_{k = 0}\sum_{i=0}^H(M_t^{[i]} - \tilde{M}_t^{[i]})w_{t-i}\| \\
    &\leq 2\kappa_B \kappa^3 (1 - \gamma)^k\PertBound \sum_{i=0}^{\hor}\left(\|M_{t-k}^{[i]} - \tilde{M}_{t-k}^{[i]}\|\right).\qedhere
\end{align*}
\end{proof}

Therefore using assumption \ref{a2} and Lemma \ref{lemma:statebounds}, we immediately get that 
\begin{align*}
    f_t(M_{t - \hor} \ldots M_{t - k} \ldots M_t) - f_t(M_{t - \hor} \ldots \tilde{M}_{t - k} \ldots M_t) \leq 2\CostGradBound\BigBound\PertBound\kappa_B \kappa^3 (1 - \gamma)^k \sum_{i=0}^{\hor}\left(\|M_{t-k}^{[i]} - \tilde{M}_{t-k}^{[i]}\|\right)
\end{align*}
\subsubsection{Bounding the Gradient}

\begin{lemma}
\label{lemma:biggradbound}
For all $M$ such that $\|M^{[j]}\| \leq a(1 - \gamma)^j $ for all $j \in [1,\hor]$,  we have that
\[\|\nabla_{M} f_t(M  \ldots M)\|_F \leq \CostGradBound \BigBound \PertBound \hor \dimn \left( \frac{2\BBound \kappa^3}{\gamma} + H\right) \]
\end{lemma}
Note that since $M$ is a matrix, the $\ell_2$ norm of the gradient $\nabla_M f_t$ corresponds to to the Frobenius norm of the $\nabla_M f_t$ matrix. Due to space constraints, we provide the proof in the appendix. 

\begin{proof}[Proof of Lemma \ref{lemma:biggradbound}]
To derive a crude bound on the quantity in question, it will be sufficient to derive an absolute value bound on $\nabla_{M^{[r]}_{p,q}} f_t(M, \ldots, M)$ for all $r,p,q$. To this end, we consider the following calculation.
Using Lemma \ref{lemma:statebounds}, we get that $y_t^K(M \ldots M), v_t^K(M \ldots M) \leq \BigBound$. Therefore, using assumption \ref{a2}, we have that
\begin{align*}
    |\nabla_{M^{[r]}_{p,q}} f_t(M \ldots M)| \leq \CostGradBound\BigBound\left(\left\|\frac{\partial y_t^K(M)}{\partial M^{[r]}_{p,q}} + \frac{\partial v_t^K(M \ldots M)}{\partial M^{[r]}_{p,q}}\right\|\right).
\end{align*}
\[ \]
We now bound the quantities on the right-hand side:
\begin{align*}
    \left\|\frac{\delta y_t^K(M \ldots M)}{\delta M^{[r]}_{p,q}}\right\| 
    &= \left\|  \sum_{i = 0}^{2\hor} \sum_{j=1}^{\hor} \left[ \frac{ \partial \tilde{A}_K^jBM^{[i-j]}}{\partial M^{[r]}_{p,q} }\right]w_{t-i} \mathbf{1}_{i-j \in [1,\hor]}\right\| \\
    &\leq \sum_{i = r}^{r + \hor}\left\| \left[ \frac{ \partial \tilde{A}_K^{i-r}BM^{[r]}}{\partial M^{[r]}_{p,q} }\right]w_{t-i}\right\| \leq \frac{\PertBound \BBound \kappa^2}{\gamma}.
 \end{align*}   
    Similarly,
    \begin{align*}
       \left\|\frac{\partial v_t^K(M \ldots M)}{\partial M^{[r]}_{p,q}}\right\|  &\leq \kappa \left\|\frac{\delta y_t^K(M \ldots M)}{\delta M^{[r]}_{p,q}}\right\| + \left\|\sum_{i=0}^{H} \frac{\partial M^{[i]}}{\partial M^{[r]}_{p,q}} w_{t-i}\right\| \\
       &\leq \PertBound\left(\frac{\BBound \kappa^3}{\gamma} + \hor\right).
    \end{align*}
Combining the above inequalities gives the bound in the lemma.\qedhere

\end{proof}

\section{Conclusion}
In this paper, we have shown how to control linear dynamical systems with adversarial disturbances through regret minimization, as well as how to handle general convex costs. Our notion of robust controller is able to learn and adapt the controller according to the noise encountered during the process. 
This deviates from the study of robust control in the framework of $H_\infty$ control, that attempts to find a control with worst-case anticipate of all future noises. 

\subsection*{Acknowledgements}
Sham Kakade acknowledges funding from the Washington Research Foundation for Innovation in Data-intensive Discovery, the DARPA award FA8650-18-2-7836, and the ONR award N00014-18-1-2247. 

\bibliography{our_bib.bib}

\begin{thebibliography}{10}

\bibitem{a2}
Yasin Abbasi{-}Yadkori, Nevena Lazic, and Csaba Szepesv{\'{a}}ri.
\newblock Regret bounds for model-free linear quadratic control.
\newblock {\em CoRR}, abs/1804.06021, 2018.

\bibitem{a1}
Yasin Abbasi{-}Yadkori and Csaba Szepesv{\'{a}}ri.
\newblock Regret bounds for the adaptive control of linear quadratic systems.
\newblock In {\em {COLT} 2011 - The 24th Annual Conference on Learning Theory,
  June 9-11, 2011, Budapest, Hungary}, pages 1--26, 2011.

\bibitem{abbasi2011regret}
Yasin Abbasi-Yadkori and Csaba Szepesv{\'a}ri.
\newblock Regret bounds for the adaptive control of linear quadratic systems.
\newblock In {\em Proceedings of the 24th Annual Conference on Learning
  Theory}, pages 1--26, 2011.

\bibitem{anava2015online}
Oren Anava, Elad Hazan, and Shie Mannor.
\newblock Online learning for adversaries with memory: price of past mistakes.
\newblock In {\em Advances in Neural Information Processing Systems}, pages
  784--792, 2015.

\bibitem{arora2012online}
Raman Arora, Ofer Dekel, and Ambuj Tewari.
\newblock Online bandit learning against an adaptive adversary: from regret to
  policy regret.
\newblock {\em arXiv preprint arXiv:1206.6400}, 2012.

\bibitem{arora2018towards}
Sanjeev Arora, Elad Hazan, Holden Lee, Karan Singh, Cyril Zhang, and Yi~Zhang.
\newblock Towards provable control for unknown linear dynamical systems.
\newblock 2018.

\bibitem{bertsekas2005dynamic}
Dimitri Bertsekas.
\newblock {\em Dynamic programming and optimal control}, volume~1.
\newblock Athena scientific Belmont, MA, 2005.

\bibitem{cesa2006prediction}
Nicolo Cesa-Bianchi and G{\'a}bor Lugosi.
\newblock {\em Prediction, learning, and games}.
\newblock Cambridge university press, 2006.

\bibitem{cohen2018online}
Alon Cohen, Avinatan Hassidim, Tomer Koren, Nevena Lazic, Yishay Mansour, and
  Kunal Talwar.
\newblock Online linear quadratic control.
\newblock {\em arXiv preprint arXiv:1806.07104}, 2018.

\bibitem{DBLP:journals/corr/abs-1805-09388}
Sarah Dean, Horia Mania, Nikolai Matni, Benjamin Recht, and Stephen Tu.
\newblock Regret bounds for robust adaptive control of the linear quadratic
  regulator.
\newblock {\em CoRR}, abs/1805.09388, 2018.

\bibitem{dekel2013better}
Ofer Dekel and Elad Hazan.
\newblock Better rates for any adversarial deterministic mdp.
\newblock In {\em International Conference on Machine Learning}, pages
  675--683, 2013.

\bibitem{even2009online}
Eyal Even-Dar, Sham~M Kakade, and Yishay Mansour.
\newblock Online markov decision processes.
\newblock {\em Mathematics of Operations Research}, 34(3):726--736, 2009.

\bibitem{fazel2018global}
Maryam Fazel, Rong Ge, Sham~M Kakade, and Mehran Mesbahi.
\newblock Global convergence of policy gradient methods for linearized control
  problems.
\newblock {\em arXiv preprint arXiv:1801.05039}, 2018.

\bibitem{hardt2016gradient}
Moritz Hardt, Tengyu Ma, and Benjamin Recht.
\newblock Gradient descent learns linear dynamical systems.
\newblock {\em arXiv preprint arXiv:1609.05191}, 2016.

\bibitem{OCObook}
Elad Hazan.
\newblock Introduction to online convex optimization.
\newblock {\em Foundations and Trends in Optimization}, 2(3-4):157--325, 2016.

\bibitem{hazan2018spectral}
Elad Hazan, Holden Lee, Karan Singh, Cyril Zhang, and Yi~Zhang.
\newblock Spectral filtering for general linear dynamical systems.
\newblock {\em arXiv preprint arXiv:1802.03981}, 2018.

\bibitem{hazan2017learning}
Elad Hazan, Karan Singh, and Cyril Zhang.
\newblock Learning linear dynamical systems via spectral filtering.
\newblock In {\em Advances in Neural Information Processing Systems}, pages
  6702--6712, 2017.

\bibitem{kalman1960new}
Rudolph~Emil Kalman.
\newblock A new approach to linear filtering and prediction problems.
\newblock {\em Journal of Basic Engineering}, 82.1:35--45, 1960.

\bibitem{ljung1998system}
Lennart Ljung.
\newblock {\em System identification: Theory for the User}.
\newblock Prentice Hall, Upper Saddle Riiver, NJ, 2 edition, 1998.

\bibitem{shalev2012online}
Shai Shalev-Shwartz et~al.
\newblock Online learning and online convex optimization.
\newblock {\em Foundations and Trends{\textregistered} in Machine Learning},
  4(2):107--194, 2012.

\bibitem{simchowitz2018learning}
Max Simchowitz, Horia Mania, Stephen Tu, Michael~I Jordan, and Benjamin Recht.
\newblock Learning without mixing: Towards a sharp analysis of linear system
  identification.
\newblock {\em arXiv preprint arXiv:1802.08334}, 2018.

\bibitem{z2}
Robert~F Stengel.
\newblock {\em Optimal control and estimation}.
\newblock Courier Corporation, 1994.

\bibitem{wang2019system}
Yuh-Shyang Wang, Nikolai Matni, and John~C Doyle.
\newblock A system level approach to controller synthesis.
\newblock {\em IEEE Transactions on Automatic Control}, 2019.

\bibitem{yu2009markov}
Jia~Yuan Yu, Shie Mannor, and Nahum Shimkin.
\newblock Markov decision processes with arbitrary reward processes.
\newblock {\em Mathematics of Operations Research}, 34(3):737--757, 2009.

\bibitem{z1}
Kemin Zhou, John~Comstock Doyle, Keith Glover, et~al.
\newblock {\em Robust and optimal control}, volume~40.
\newblock Prentice hall New Jersey, 1996.

\end{thebibliography}
\bibliographystyle{plain}

\appendix
\section{Appendix}

\section{Proof of Theorem \ref{thm:oco_memory}}
\begin{proof}
By the standard OGD analysis, we know that
\begin{equation*}
    \sum\limits_{t=H}^T \tilde{f}_t(x_t) - \min\limits_{x \in \K} \sum\limits_{t=H}^T \tilde{f}_t(x) \leq \frac{D^2}{\eta} + TG^2\eta.
\end{equation*}
In addition, we know by \eqref{eq:memlip} that, for any $t \geq H$,
\begin{align*}
\abs{f_t(x_{t-H},\dots,x_t) - f_t(x_t,\dots,x_t)} &\leq L\sum\limits_{j=1}^H\norm{x_t - x_{t-j}} \leq L\sum\limits_{j=1}^H\sum\limits_{l=1}^j\norm{x_{t-l+1} - x_{t-l}}\\
&\leq L\sum\limits_{j=1}^H\sum\limits_{l=1}^j \eta\norm{\grad \tilde{f}_{t-l}(x_{t-l})} \leq LH^2\eta G,
\end{align*}
and so we have that
\begin{equation*}
\abs{\sum\limits_{t=H}^T f_t(x_{t-H},\dots,x_t) - \sum\limits_{t=H}^T f_t(x_t,\dots,x_t)} \leq TLH^2\eta G.
\end{equation*}
It follows that
\begin{align*}
    \sum\limits_{t=H}^T f_t(x_{t-H},\dots,x_t) &- \min\limits_{x \in \K} \sum\limits_{t=H}^T f_t(x,\dots,x) \leq \frac{\memdiam^2}{\eta} + T\memgradbound^2\eta + LH^2\eta \memgradbound T.
\end{align*}\qedhere
\end{proof}

\begin{algorithm}[t!]
\caption{OGD with Memory (OGD-M).}
\label{ogdm}
\begin{algorithmic}[1]
\STATE \textbf{Input:} Step size $\eta$, functions $\braces{f_t}_{t=m}^T$
\STATE Initialize $x_0, \dots, x_{\hor-1} \in \K$ arbitrarily.
\FOR{$t = H, \ldots, T$}
\STATE Play $x_t$, suffer loss $f_t(x_{t-H},\dots,x_t)$
\STATE Set $x_{t+1} = \Pi_{\K}\pa{x_t - \eta\grad \tilde{f}_t(x)}$
\ENDFOR
\end{algorithmic}
\end{algorithm}


\end{document}